\documentclass{anthology-ch}         

\usepackage{booktabs}
\usepackage{graphicx}

\title{"Too Much Alignment; Not Enough Culture": Re-balancing cultural alignment practices in LLMs}

\author[1]{Eric J. W. Orlowski}[
  orcid=0000-0002-4022-9948
]
\author[2]{Hakim Norhashim}[orcid=0009-0009-1578-653X]
\author[3]{Tristan Koh Ly Wey}[orcid=0009-0001-9602-0844]

\affiliation{1,2,3}{AI Singapore, National University of Singapore}


\pubyear{2025}
\pubvolume{1}
\pagestart{1}
\pageend{1}
\conferencename{Proceedings of Conference XXX}
\conferenceeditors{Editor1 Editor2}
\doi{00000/00000}  

\begin{document}

\maketitle

\begin{abstract}
While cultural alignment has increasingly become a focal point within AI research, current approaches relying predominantly on quantitative benchmarks and simplistic proxies fail to capture the deeply nuanced and context-dependent nature of human cultures. Existing alignment practices typically reduce culture to static demographic categories or superficial cultural facts, thereby sidestepping critical questions about what it truly means to be culturally aligned. This paper argues for a fundamental shift towards integrating interpretive qualitative approaches drawn from social sciences into AI alignment practices, specifically in the context of Large Language Models (LLMs). Drawing inspiration from Clifford Geertz’s concept of “thick description,” we propose that AI systems must produce outputs that reflect deeper cultural meanings—what we term “thick outputs”—grounded firmly in user-provided context and intent. We outline three necessary conditions for successful cultural alignment: sufficiently scoped cultural representations, the capacity for nuanced outputs, and the anchoring of outputs in the cultural contexts implied within prompts. Finally, we call for cross-disciplinary collaboration and the adoption of qualitative, ethnographic evaluation methods as vital steps toward developing AI systems that are genuinely culturally sensitive, ethically responsible, and reflective of human complexity.
\end{abstract}

\section{Introduction} 

Cultural alignment is a core challenge within AI development. The drive to align models with culture–often through some proxy like cultural values \cite{adilazuarda-etal-2024-towards}–is not only intrinsically valuable to produce fairer AI systems, but also remains instrumentally valuable, as one instance of a broader class of domains that are inherently fractal in nature. Its instrumentality lies in that solving these problems brings us closer to meeting other, similarly structured, challenges. One view of cultural alignment involves “tailoring AI systems to mirror the collective beliefs, values, and norms of specific user groups to enhance user interactions with AI and prevent cultural misunderstandings” \cite{masoud2025cultural}. Alternatively, a culturally aligned model is one that reflects aspects of life (i.e. knowledge, belief, art, morals, law, custom and any other capabilities and habits) that are aligned with a group of people \cite{alkhamissi-etal-2024-investigating}, \cite{romero2024cvqa}. From the Natural Language Processing (NLP) perspective, the main intersections of culture with language include linguistic form and style, common-ground knowledge, aboutness and objectives/values \cite{hershcovich-etal-2022-challenges}.

These broad definitions of cultural alignment are delineated into schemas using demographic proxies such as nationality, ethnicity, language, or values \cite{adilazuarda-etal-2024-towards}. Examples include defining culture as various collections of facts \cite{qiu2024evaluating}, e.g. values (‘Singaporeans value being on time’); domain-specific facts (‘where in Singapore is the Merlion located?’); or the level of preference with value-laden questions (‘Do you agree that on the whole, men make better business executives than women do’) adapted from surveys like the Value Survey Modules, World Values Survey, Pew Values Survey \cite{durmus2023towards}, \cite{li2024culturellm}, \cite{masoud2025cultural}. These schemas are then used to structure, collect and label data for model fine-tuning or evaluation, and translated into technical methods for cultural alignment such as benchmarking methods \cite{cahyawijaya2024cendol}, measurement metrics \cite{sukiennik2025evaluation}, prompt strategies \cite{alkhamissi-etal-2024-investigating}, \cite{tao2024cultural}, custom data collection pipelines \cite{wibowo2023copal}, human data annotation guidelines \cite{wibowo2023copal}, or prompt tuning \cite{masoud2025cultural}.

However, before these definitions of culture and schemas are operationalised into technical alignment methods and ever increasingly large datasets, they should first be critically examined and clearly justified \cite{engdahl2024agreements}. However, such definitions of culture, schemas and cultural proxies are usually assumed to be culturally relevant, and are operationalised into data collection schemas and technical methods. While seemingly straightforward, this data-driven, checklist approach simplifies culture into static stereotypes, and cultural alignment, an ever-increasing accumulation of ‘cultural data’, neglects the fluid, context-dependent, and deeply nuanced nature of human cultures. In addition, such methods reflect a familiar pattern from the standard Machine Learning playbook: avoiding human-defined concepts while letting the AI ‘learn’ from data. 

In the case of cultural alignment, the expectation is that machines can somehow understand and interpret this deeply human concept solely from data and apply it meaningfully for users. Meanwhile, humans cannot fully interpret what the machine means by “cultural alignment” (see: \cite{burrell2017fieldsite}).

The implications are many:

\begin{enumerate}
    \item \textbf{Culture is implicitly treated as the sum of datapoints, without explicit or clear definitions.}

    Implicit in this approach is the assumption that cultural alignment will emerge naturally from the accumulation of loosely defined “cultural data,” without any need to explicitly articulate what a model is being aligned towards. As a result, AI developers avoid explicitly defining what culturally meaningful outputs should look like.

    \item \textbf{AI defines what culturally aligned outputs are.}

    As a result, cultural “alignment” ends up being defined by models. Models learn from data that may be vast but uneven, unrepresentative, and collected without theoretical guidance. This makes fine-tuning efforts more inconsistent than a deliberate process informed by human expertise.

    \item \textbf{Human expertise is sidelined in an inherently human field of study.}

    Cultural alignment is rooted in lived experience and represents a form of embodied knowledge. These are domains long studied by social scientists. Current approaches treat culture as data, stripping it of the context and meta-knowledge that shape its meaning, while ignoring theoretical frameworks that provide interpretative depth.
\end{enumerate}

Our work proposes a significant shift from these conventional methods towards an interpretive, qualitative approach grounded in the social sciences. Drawing inspiration from Clifford Geertz's thick description \cite{clifford1973thick}, our approach emphasises that LLMs should produce outputs that can capture deeper meanings and contexts rather than simply articulating cultural artefacts in a superficial manner. Implicit in doing so is the need for AI systems to adopt a richer and more reflective understanding of culture in order to navigate its human interactions effectively and ethically.

\section{The Challenge of Defining Culture}

Defining ‘culture’ continues to be debated among social scientists \cite{kroeber1952culture}, \cite{yengoyan1986theory}, \cite{baldwin2006redefining}, \cite{causadias2020culture}, \cite{santos2024culture}. Most definitions refer to culture as a “collective programming of the mind” \cite{hofstede2001culture}, or “shared set of [...] values, ideas and concepts” \cite{hudelson2004culture}, or a “system of people, places and practices” \cite{causadias2020culture} which function to set one group apart from another. 

However, presenting culture as an overarching system of explicit and implicit rules for social organisation and meaning-creation is often far too general and abstract to be easily operationalised by other disciplines, such as AI developers and professionals. As Causadas summarises: “unlike crisp concepts that have relatively fixed meaning, set properties, and stable boundaries; fuzzy concepts have many layers of significance, changing their meanings according to situations [...]. This fuzziness is a main reason why it is so difficult to define culture” \cite{causadias2020culture}. Specifically, culture’s inherently context-specific nature makes it especially challenging to operationalise.

A crucial point of consideration is that humans struggle to explicitly and systematically articulate these cultural nuances, instead relying on an inherent instinctive or intuitive understanding of our cultural systems. This raises questions about what might realistically be expected from contemporary AI systems. After all, if culture is something humans largely internalise, and perhaps even something emergent, how can anyone expect machines, whose learning is dependent on context-poor traces of culture embedded in unstructured data, to navigate this complexity in any meaningfully ‘human’ sense? It may be tempting to jettison the idea altogether, as potentially unworkable, to be replaced with various proxies to (largely) fill the void \cite{adilazuarda-etal-2024-towards}—but we wish to present alternatives in this piece.

\section{What Does It Mean To Align With Culture?}

Human-AI alignment has been a central concern across industry, regulatory, and academic communities \cite{IMDA}, \cite{shen2024towards}, \cite{huang2025values}. Though initially framed by philosophers such as Nick Bostrom \cite{mulgan2016superintelligence} and AI researchers like Stuart Russell \cite{russell2022human}, the notion of alignment has since evolved into a more operational concept. Today, alignment is typically framed around three key goals:

\begin{enumerate}
    \item Accuracy, especially in domains like law and medicine, where factual correctness is critical \cite{dahl2024large}, \cite{OpenAI}, \cite{peters2025generalization};
    \item Reflection of human values, as articulated in numerous AI ethics frameworks \cite{jobin2019global}; and
    \item Avoidance of harm, particularly in the context of AI safety, security and misuse \cite{AIActionSummit}.
\end{enumerate}

Underlying all of these is a deeper sentiment: that our AI systems should reflect human intentions. Yet as AI development increasingly leans on trial-and-error, benchmark-driven performance, and optimisation without theory, our ability to trace or define those intentions is diminishing. This contributes to what is often called the black-box problem \cite{bathaee2017artificial}, \cite{burrell2016machine}.

Although significant progress has been made in aligning AI systems with respect to accuracy, ethical principles, and harm reduction, progress remains elusive in the domain of cultural alignment. Perhaps much of it has to do with the fact that cultural alignment is fundamentally different from the three categories of alignment mentioned above. Unlike accuracy or safety, which can be benchmarked, or ethical principles, which are often codified into high-level guidelines, culture, with its myriad definitions even amongst anthropologists, is inherently elusive, as discussed above. Moreover, because culture is deeply contextual, it is also interconnected and contingent, making it resistant to simplistic abstractions.  As a result, it cannot be easily captured by the types of frameworks or schemas typically used in conventional alignment approaches.
 
More specifically for language models, cultural alignment lies beyond both the syntax and direct semantics of textual data that LLMs are trained on. Culture is not just words, but lives in context, subtext, and subjective experiences. It involves deeply embedded practices. Thus, there is an epistemological gap that precludes LLMs from being able to learn cultural alignment simply from the unguided evolution through the epochs of deep learning algorithms alone.
 
The solution, in our view, is to be intentional: to surface the implicit, to convert the embodied into structured signals where possible for the algorithms to pick up, and to treat cultural nuance not as noise in the data but as a core alignment target. Culture has, for decades, been studied in situ: embedded in the contexts where it is lived and made meaningful. Cultural alignment, if it is to be taken seriously, should be approached in the same way.

\section{Why Current Approaches Fall Short}

Current approaches to AI cultural alignment are attractive because they lend themselves to measurable outcomes: either the AI model gets it right or it does not, and because models can be benchmarked and compared. However, such binary or quantitative measures fail to account for the inherent complexity and depth of culture.

As mentioned in a previous section, culture is not merely a set of static attributes but something that people continuously enact and experience through daily interactions and relationships. Thus, by definition, it cannot be entirely captured by simple factual statements or demographic categories as practised in current data collection approaches. The challenge is further compounded by the fact that current data collection approaches fail to align intentions with outcomes, despite claims to the contrary.
Current cultural data collection efforts typically rely on static categories, such as race, religion, or nationality, under the assumption that these serve as valid proxies for culture (e.g. \cite{adilazuarda-etal-2024-towards}). This creates two critical gaps. First, as discussed previously, it delegates the definition of cultural alignment to the model itself. Second, it reduces the notion of culturally aligned outputs to those that merely resemble a predefined set of cultural information embedded in external benchmarks.
These benchmarks are often drawn from independent studies or compiled lists of cultural ‘facts’ and assumptions (e.g. values and landmarks). They may not reflect the model’s training or fine-tuning data, nor are they grounded in a shared understanding of what cultural alignment should entail. This process thus implicitly defines cultural alignment of outputs as resemblance to benchmarks without clarifying why these benchmarks are meaningful to begin with. As a result, both the intentions behind cultural alignment, and the benchmarks used to evaluate it remain undefined from our perspective, and are therefore inevitably misaligned.

Moreover, culture is a form of embodied knowledge, presenting additional challenges for models built predominantly on language. While multi- and omnimodal models incorporating audio or video may offer new ways of capturing culture in context and producing more situated outputs, this does not remove the need for intentionality. The data collection process must still be curated to reflect the relevant context, and efforts must be made to surface that context where necessary to support the model’s learning. Crucially, the intention behind cultural alignment, its working definition, and the criteria for what constitutes a culturally aligned outcome must be clearly established from the outset.

\section{The Fractal Complexity of Culture}

Cultural complexity can be likened to a fractal: conceptually, culture remains effectively infinitely complex the more one zooms in or zooms out; it exhibits complex and intricate detail at every scale. This is not in itself a novel observation. However, it does mean that trying to culturally align an AI model across all possible cultural dimensions simultaneously is akin to attempting to capture infinite complexity. It further means that this is not merely a question of gathering more data to feed into a model. Rather, practical cultural alignment demands a clear delineation of specific cultural contexts that the AI model aims to reflect.

This approach is already common among the social sciences. These disciplines routinely and effectively negotiate these intellectual impasses, specifically by using targeted definitions of culture. In other words, ‘culture’ writ large is scaled down into discrete ‘chunks’, or cultural contexts \cite{agar2006ethnography}, \cite{burrell2017fieldsite}. By approaching cultural alignment along similar lines, it opens up a new path forward. Though focusing on cultural context may appear more limiting, we argue this is not the case. Indeed, the goal for AI is not perfect cultural fluency, but sufficient alignment to navigate social contexts comparably to humans.

Rather than pursuing broad cultural alignment, which quickly becomes overwhelming and ineffective, we advocate a targeted approach. Each specific context provides clearly defined parameters that make effective cultural reflection feasible and meaningful. The core insight of our argument is the recognition that AI can thus never fully ‘represent’ culture in a comprehensive sense. While representation in some form is inherent to a model, representation also implies completeness and authority, which, we argue, with regard to culture should be beyond the capability of current AI models. 

An analogy might be useful here: imagine translating a three-dimensional globe onto a two-dimensional map. Despite advances in mapping technology, distortions are unavoidable. Similarly, AI models translating human cultural nuances into text-based outputs inevitably produce distortions. Accepting this limitation allows developers to shift their focus towards creating thick outputs rather than (likely) unattainable perfect representations.

\section{Producing Thick Outputs}

Crucially, our framework necessitates that AI models generate what we call thick outputs. Borrowing Geertz’s idea of thick description \cite{clifford1973thick}, thick outputs refer to responses from AI that encompass deeper cultural meanings, not merely surface-level correctness. Geertz famously illustrated this by distinguishing between a twitch (an involuntary eyelid movement) and a wink (a deliberate, communicative gesture). Mechanically, these actions are identical, but culturally, they are profoundly different.

Applying this concept to AI, a thick output would enable an AI system to distinguish nuanced, context-specific cultural differences similarly. For instance, a culturally aligned AI chatbot interacting with users from different cultural backgrounds would not only translate their language accurately but also reflect cultural context, nuances of formality, and implicit meanings that convey respect, hierarchy, or social relationships.

In addition, an essential but often overlooked element of cultural alignment is the relationship between a model’s output and the prompt it receives. For a model to reflect cultural elements in a way that enables users to attribute meaningful cultural interpretation, the output must not only be ‘thick’ but must also be anchored in the cultural cues and intent implied within the user’s prompt.

This anchoring is crucial. Given the fractal and contingent nature of culture, even a model with an excellent internal representation of a culture and the capacity to produce thick outputs cannot guarantee cultural alignment. Alignment depends on the model's ability to respond contextually to a specific prompt: one embedded with cultural assumptions, framing, or positionality. Without that connection, users may be unable to interpret the output as culturally meaningful.

From this, we derive three independent necessary conditions for successful cultural alignment:

\begin{enumerate}
    \item \textbf{Cultural representation:}
    As mentioned in previous sections, the model must contain a sufficiently scoped internal representation of the cultural context it seeks to align with, bounded by the intended use or application domain.

    \item \textbf{Capacity for Thick Outputs:}
    The model must be capable of generating outputs with layered cultural nuance, which we have previously defined as ‘thick outputs’.

    \item \textbf{Prompt Anchoring, resulting in Reflection:}
    The outputs must reflect the user’s context and intent, as expressed or implied in the input prompt and the surrounding interaction. Without this anchoring, alignment cannot occur in a meaningful way.
\end{enumerate}

Taken together, these conditions suggest that cultural alignment is not a generalisable, model-level property; it is only meaningful at the level of individual interactions. It is users, not models, who ultimately attribute cultural meaning to outputs. 

This has critical implications for evaluation: most benchmarks assume a one-size-fits-all relationship between prompts and responses. But cultural alignment demands situational specificity. A prompt that elicits a culturally appropriate output in one context may not do so in another, making general-purpose benchmarking insufficient for assessing alignment.

A significant challenge facing cultural alignment in AI is evaluating whether an AI is genuinely culturally sensitive and reflective. Conventional quantitative benchmarks, such as accuracy scores, are insufficient for capturing cultural complexity. A model scoring 90\% accuracy on culturally aligned questions does not necessarily mean it adequately reflects cultural depth and nuance.

We advocate using qualitative and ethnographic research methods to develop evaluations with greater validity. Such methods could include user feedback from culturally diverse groups, ethnographic analysis of AI interactions in real-world scenarios, and iterative testing of AI-generated responses. This approach provides richer insights into how well AI systems reflect the intended cultural contexts, thus guiding more effective model development and deployment.

\section{Conclusions}

We have argued that current approaches to cultural alignment, primarily within the context of LLMs, can be improved by adopting an approach balancing the technical alignment processes with clear and earnest engagement with socially scientific perspectives on culture. Currently, too much focus is placed on technical alignment of LLMs, with insufficient emphasis on clarifying precisely what the model is aligned with. We appreciate that this is a big challenge in its own right; social scientists long conceptualised culture differently. Advocating for a more balanced approach invites additional challenges by having to translate a multiplicitous and interdisciplinary field with limited consensus not just to another field (computer science and AI engineering), but to also find effective ways to operationalise these concepts.

We advocate for (1) clearly defined cultural contexts and (2) high-quality, context-specific data. We also recognise the need to develop new workflows and best practices—e.g., reinforcement learning with human feedback (RLHF)—and developing benchmarks and other quantitative and qualitative validation frameworks. Such steps are by no means trivial. Achieving meaningful cultural reflection in AI systems therefore cannot be the domain of one or a few disciplines. To successfully meet such challenges requires close collaboration across disciplinary lines. Developing culturally reflective AI requires careful, ethical data collection and training. Collaboration between technical and social scientists ensures these considerations remain central. Ultimately, our approach seeks to move beyond the limited paradigm of cultural representation towards a richer, more culturally reflective AI. By embracing the inherent complexity and depth of human cultures, our framework aims to create AI systems that are not just technically proficient but also ethically responsible and culturally sensitive.



\printbibliography

\end{document}